\documentclass[letterpaper, 10 pt, journal, twoside]{IEEEtran}

\usepackage{graphicx}
\usepackage{amsmath}
\usepackage{algorithm}
\usepackage{algpseudocode}
\usepackage{xcolor}
\usepackage{soul}
\usepackage{stfloats}
\usepackage{amsfonts} 
\usepackage{amssymb}
\usepackage{tabularx}
\usepackage{booktabs}
\usepackage{comment}
\usepackage{subcaption}
\renewcommand{\baselinestretch}{1}
\usepackage{xcolor}
\usepackage{comment}
\usepackage{hyperref}
\newcommand\edit[1]{{\color{black}#1}}
\usepackage{indentfirst}
\usepackage{tkz-graph}
\usetikzlibrary{automata,positioning}
\usepackage{tikz}
\usetikzlibrary{positioning}
\usepackage[nolist,nohyperlinks]{acronym}

\def\identitymat{\mathbf{I}}

\def\occ{o}
\def\distfield{{\hat{d}}}

\def\revert{r}

\def\lengthscale{l}

\def\abscissavec{\mathbf{x}}
\def\abscissamat{\mathbf{X}}

\newcommand{\kernelraw}[1]{{{k}_{#1}}}

\newcommand{\kernelmatraw}[1]{{\mathbf{K}_{#1}}}

\newcommand{\kernelvecraw}[1]{{\mathbf{k}_{#1}}}

\ifCLASSINFOpdf
\else
\fi

\begin{document}

\title{Interactive Distance Field Mapping and Planning to Enable Human-Robot Collaboration}


\author{Usama Ali$^{1*}$, Lan Wu$^{2*}$, Adrian M\"uller$^{1}$, Fouad Sukkar$^{2}$, Tobias Kaupp$^{1}$, Teresa Vidal-Calleja$^{2}$
\thanks{This work has been submitted to the IEEE for possible publication. Copyright may be transferred without notice, after which this version may no longer be accessible.}
\thanks{This work was supported by the Industrial Transformation Training Centre (ITTC) for Collaborative Robotics in Advanced Manufacturing (also known as the Australian Cobotics Centre) funded by ARC (Project ID: IC200100001) and the Bavarian Research Foundation (grant AZ-1512-21).} 
\thanks{$^{*}$These authors are co-first authors and contributed equally to this work. Corresponding author: {\tt\footnotesize Lan.Wu-2@uts.edu.au}}
\thanks{$^{1}$Authors are with the Center for Robotics (CERI) at the Technical University of Applied Sciences W\"urzburg-Schweinfurt (THWS), Germany.}
\thanks{$^{2}$Authors are with the Robotics Institute, Faculty of Engineering and IT, University of Technology Sydney (UTS), Australia.}
}



\maketitle

\begin{abstract} 
Human-robot collaborative applications require scene representations that are kept up-to-date and facilitate safe motions in dynamic scenes. In this letter, we present an interactive distance field mapping and planning (IDMP) framework that handles dynamic objects and collision avoidance through an efficient representation. We define \textit{interactive} mapping and planning as the process of creating and updating the representation of the scene online while simultaneously planning and adapting the robot's actions based on that representation.  
The key aspect of this work is an efficient Gaussian Process field that performs incremental updates and handles \edit{dynamic objects reliably by identifying moving points via} a simple and elegant formulation based on queries from a temporary latent model. In terms of mapping, IDMP is able to fuse point cloud data from single and multiple sensors, query the free space at any spatial resolution, and deal with moving objects without semantics. In terms of planning, IDMP allows seamless integration with gradient-based \edit{reactive planners facilitating dynamic obstacle avoidance for safe human-robot interactions}. Our mapping performance is evaluated on both real and synthetic datasets. A comparison with similar state-of-the-art frameworks shows superior performance when handling dynamic objects and comparable or better performance in the accuracy of the computed distance and gradient field. Finally, we show how the framework can be used for fast motion planning in the presence of moving objects~\edit{both in simulated and real-world scenes}. An accompanying video, code, and datasets are made publicly available\footnote[3]{\tt \url{https://uts-ri.github.io/IDMP}}.
\end{abstract} 


\IEEEpeerreviewmaketitle

\section{Introduction}

Human-robot
collaboration (HRC) and other applications of robots in the field call for interactive representations to deal with dynamic and evolving scenes. For true collaboration in industrial settings, humans and robots physically share the same space, e.g. working jointly and simultaneously on the assembly of a product. The environment consisting of the human, robot, workspace, and objects such as tools and assembly components needs to be monitored by sensors and kept up-to-date in a timely manner. 
One approach is to create and continuously update an interactive representation of the changing scene, which is useful for safe and effective robot motion planning around human operators.

\begin{figure}[t]
    \centering
    \includegraphics[width=0.95\linewidth]{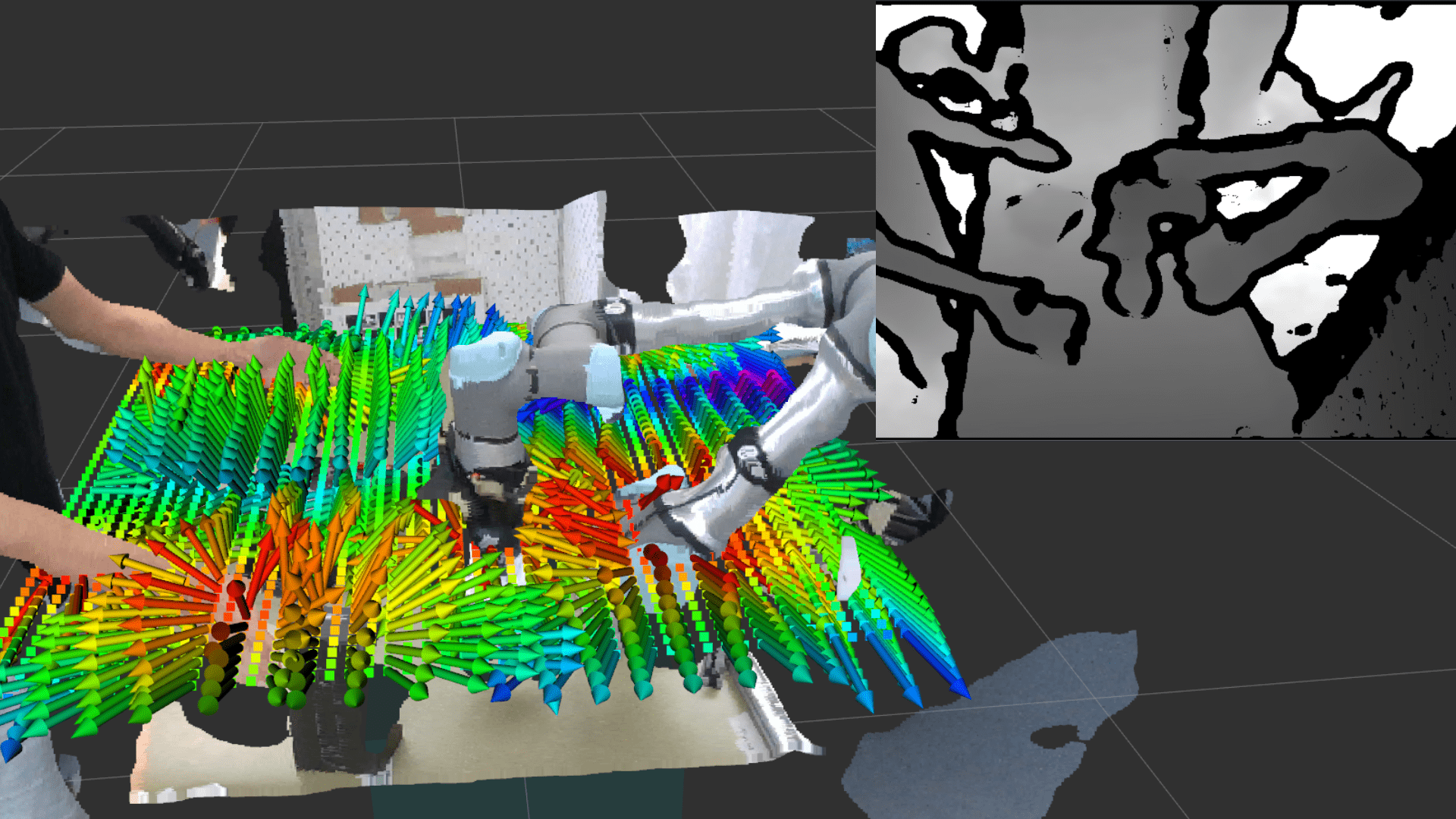}
    \caption{Interactive generation of a distance and gradient field in an HRC setting. Top right: Depth image from an Intel Realsense camera. Bottom left: Coloured point cloud and a horizontal slice of the field (red/blue means close/far to the nearest object with arrows pointing away from it).}
    \vspace{-2ex}\label{fig:dist_field_example_HRC}
\end{figure}

Recent representations that aim to fulfil some of the above-mentioned requirements have been proposed in the robotics literature~\cite{han_fiesta_2019,bai_vdbblox_2023}. \ac{EDF}, for instance, has recently become a promising representation suitable for direct collision-checking. As an example, the methods in~\cite{oleynikova_voxblox_2017,pan_voxfield_2022,bai_vdbblox_2023} compute the projective or non-projective \ac{TSDF} first, then propagate from \ac{TSDF} to the free space for updating the EDF. Han et al.~\cite{han_fiesta_2019} propose to propagate from an occupancy map instead of the TSDF to maintain efficiency while preserving accuracy for the EDF computation. EDF representations also have the ability to dynamically update the changes via the so-called free space carving method. This method performs expensive ray-casting and progressive integration of the TSDF to update the map, resulting in free space that is only gradually cleared when objects move in the scene.

This paper presents an interactive distance field mapping and planning (IDMP) framework aimed at dynamic scenes common in human-robot collaboration scenarios (see Fig.~\ref{fig:dist_field_example_HRC} as an example). \edit{By extending our previous work in~\cite{wu_faithful_2021,gentil_accurate_2023}, we focus to build and maintain an efficient and up-to-date distance and gradient field} using a Gaussian-Process-based method capable of integrating depth sensor measurements.  
\edit{Here, we present a simple and elegant method catering both dynamic updates and seamless fusion} by querying distances and gradients from a temporary latent GP distance field generated with only the points of the current frame. We name this latent representation the ``Frustum Field". The Frustum Field is able to deal with the discrepancy in the distance between surface points that move in the current frame without the need for ray tracing. It also allows fusion with previously mapped distance and gradient fields.

To overcome the computational complexity of GP-based approaches, our framework maintains the updates and fusion in an Octree structure that runs online. Furthermore, our method handles dynamic objects reliably via the information provided by the Frustum Field. 
The fused representation is continuous, allowing querying of the Euclidean distance to the nearest surface and its gradient at an arbitrary spatial resolution. 
\edit{To demonstrate the potential of our efficient and differentiable fused distance and gradient fields for human-robot interaction tasks, we integrate IDMP with a gradient-based motion planner as well as a local reactive behaviour.
}
We experimentally evaluate IDMP's performance in both static and dynamic scenes and benchmark it against state-of-the-art algorithms \edit{and demonstrate its capabilities in real-world experiments for a robot arm interacting with a human}. 

To summarise, the key contributions of the  paper are:
\begin{itemize}
\item A novel dynamic update and fusion method for GP-based distance and gradient fields. This method is based on a latent representation named Frustum Field that enables dealing with static, moving, and new points.
\item An interactive distance field mapping and planning (IDMP) framework that generates a continuous Euclidean distance and gradient field online, and is integrated with gradient-based \edit{reactive motion and path planner for 3D collision-free interactions}.  
\item 
An efficient open-sourced ROS-based implementation of the full IDMP framework.~\footnote{Code is made publicly available at \url{https://github.com/UTS-RI/IDMP}.}
\end{itemize}

To the best of our knowledge, this work is the first framework that addresses the close interactivity between perception and motion planning aimed at HRC applications.


\section{Related Work}\label{sec:related_work}


\ac{EDF}s are useful representations for a wide range of tasks which require distance-to-obstacle information at different points in space. For example, motion planners such as gradient-based trajectory optimisers~\cite{zucker2013chomp, mukadam2018continuous} utilise EDFs and their gradients to perturb trajectories smoothly in order to avoid obstacles. EDFs can be computed by solving the Eikonal equation. A major challenge in exploiting the Eikonal equation to recover the ESDF is that it is non-linear and hyperbolic. Therefore, it is difficult to solve directly, and specialised numerical approaches are often required, including discrete methods such as fast marching or label-correcting to propagate the distance field through a grid. 

Voxblox~\cite{oleynikova_voxblox_2017} proposes an approach to compute the ESDF based on the projective TSDF through a wavefront propagation algorithm. Voxfield~\cite{pan_voxfield_2022} and VDBblox~\cite{bai_vdbblox_2023} adopt the use of a non-projective TSDF to improve the accuracy. FIESTA~\cite{han_fiesta_2019} proposes to use an occupancy map instead of TSDF and a novel data structure to maintain efficiency. VED-EDT uses a distance transform function to represent the EDF hierarchically~\cite{zhu_vdb-edt_2021}. These EDF representations both have the capability to dynamically update the changes via the so-called free space carving method, which updates every voxel along the grouped rays. This method requires performing expensive ray-casting and progressive integration and weighting of the TSDF to update the map, resulting in free space that is only gradually cleared when objects move in the scene. There are also frameworks that explicitly deal with dynamic objects~\cite{schmid2023dynablox}. However, the dynamic object handling is decoupled from mapping. Our framework deals reliably with dynamic objects and their surfaces are considered for collision avoidance, thus tightly coupled with the estimated representation.

Over the last few years, there have been numerous works exploring the potential of learning approaches such as neural networks for implicit signed distance representations. In particular, inspired by~\cite{park_deepsdf_2019}, the authors in~\cite{gropp_learning_shapes_2020} propose to use an implicit neural representation to constrain the learning to approximate the SDF. Recently proposed in~\cite{pantic_NeRF_planning_2022}, the authors learn an approximation of EDF from Neural Radiance Fields (NeRFs) with occupancy inference. Similarly, inspired from~\cite{gropp_learning_shapes_2020}, iSDF~\cite{ortiz_isdf_2022} is proposed to use a neural signed distance field for mapping. For dynamic scenes, D-NeRF considers time as an additional input to the system and splits the learning process into two main stages: one that encodes the scene into a canonical space and another that maps this canonical representation into the deformed scene at a particular time. While these approaches have benefits such as a continuous representation and lower memory cost, they are expensive to compute and require powerful GPUs.

GP-based distance fields are appealing as they can represent complex environments with non-parametric models~\cite{bhoram_online_2019,wu_faithful_2021}. In our previous work, we produce the continuous distance field by formulating the Gaussian Process via a latent distance function. The regularised Eikonal equation can be simply solved by applying the logarithmic transformation to a GP formulation to approximate the accurate Euclidean distance fields~\cite{wu_faithful_2021,wu_log-gpis-mop_2023}. In \cite{wu_log-gpis-mop_2023} a trajectory optimisation approach for motion planning is implemented for the 2D case. In this work, we generalise and apply this approach to 3D. 

Following the theoretical formulation in~\cite{wu_faithful_2021,wu_log-gpis-mop_2023}, we have introduced the so-called reverting GP distance field~\cite{gentil_accurate_2023} that increased the accuracy of the distance field addressing multiple key problems in robotics with noisy measurements. A GP distance field has been applied to the active dynamic mapping framework in~\cite{liu_active_2021} to enable accurate distance inferences~\cite{wu_pseudo_2023}. However, ~\cite{liu_active_2021,wu_pseudo_2023} requires the inverse depth to formulate the occupancy field for dynamic updates. In contrast, we present a simple and elegant method for dynamic updates and fusion of our GP distance field by querying distance and gradients from a latent GP distance field generated with only the points of the current frame. Our proposed framework and its implementation is faster than all our previous works, making it suitable for close-proximity HRC tasks.
In the context of HRC, it is important to have an efficient representation that informs the planner of a possible collision, in particular, when objects are moving in the scene.
Two recent review papers discuss the state of the art in safe HRC~\cite{robla-gomez_working_2017, patil_advances_2023}. Both distinguish between methods that are aimed at contact detection/mitigation, and collision avoidance. Our approach follows the latter which is enabled by the interactivity between mapping and motion planning.

Many obstacle avoidance methods for HRC explicitly track the human operator using RGB-D cameras~\cite{zanchettin_safety_2016, schirmer2023anomaly,morato_toward_2014}. Human limbs are represented using skeletons~\cite{schirmer2023anomaly,morato_toward_2014}, bounding spheres~\cite{morato_toward_2014}, cylinders~\cite{schirmer_towards_2024} or an occupancy box~\cite{tonola2021anytime}. Based on the model of the human and the known state of the robot, the control strategy is often based on discrete modes, e.g. "autonomous behaviour"~\cite{zanchettin_safety_2016}. In some HRC control frameworks, the human position is assumed to be known, e.g. \cite{faroni_mpc_2019}. In contrast to those approaches, our framework does not rely on semantics enabling workspace monitoring of all types of objects (static and dynamic) using a continuous field representation. Our efficient map update allows an online use of standard trajectory optimisation algorithms rather than discrete control modes. Another non-semantic approach is given in \cite{flacco_depth_2015, flacco_real-time_2017}. However, they only compute distances to discrete control points while our approach is able to recover a continuous field including non-observed areas.



\section{Background}
\label{sec:background}
\subsection{Gaussian Process Distance Field}

Gaussian Process (GP)~\cite{GPbook} is a flexible and probabilistic approach to modelling distributions over functions, particularly suitable in scenarios with noisy and sparse measurements.
Let us consider a set of noisy input observations (training points) $\mathbf{y}=\left\{y_{j} = f(\mathbf{x}_{j}) + \epsilon_{j} \right\}_{j=1}^{J} \in \mathbb{R}$ taken at locations $\mathbf{X}=\{\mathbf{x}_j\}_{j=1}^{J} \in \mathbb{R}^{D} $ and corrupted by additive Gaussian noise $\epsilon_{j} \sim \mathcal{N}(0, \sigma^2)$. We model the unknown distribution $f$ as a GP:
\begin{equation}
    f \sim \mathcal{GP}(0, k_{\mathbf{x}\mathbf{x}'}),
\end{equation} 
    where $k_{\mathbf{x}\mathbf{x}'}$ is the covariance function that controls the properties of $f$. Suppose we have a set of testing points $\mathbf{X_*}=\{\mathbf{x_*}_q\}_{q=1}^{Q} \in \mathbb{R}^{D} $. At the testing locations, we can express the joint distribution of the function values and the observed target values using the kernel function,
\begin{equation}\label{prior_GP}
\left[\begin{array}{l}
\mathbf{y} \\
\mathbf{f}_*
\end{array}\right]=\mathcal{N}\left(\mathbf{0},\left[\begin{array}{ll}
\mathbf{K}_{\mathbf{X} \mathbf{X}}+\sigma_{f}^2 \mathbf{I} & \mathbf{K}_{\mathbf{X} \mathbf{X_*}} \\
\mathbf{K}_{\mathbf{X_*} \mathbf{X}} & \mathbf{K}_{\mathbf{X_*} \mathbf{X_*}}
\end{array}\right]\right).
\end{equation}
$\mathbf{f}_*$ is the vector of predicted function values at $\mathbf{X_*}$, $\mathbf{I}$ is the identity matrix, $\sigma_{f}^2$ is the variance of observation noise and $\mathbf{K}_{\mathbf{X} \mathbf{X}}$ is the $J \times J$ covariance matrix between input points. By conditioning Eq.~\eqref{prior_GP}, the posterior distribution of $f$ at an arbitrary testing point $\mathbf{x}_{*}$ is given by $f(\mathbf{x}_{*}) \sim \mathcal{N}(\hat{f}\left(\mathbf{x_*}\right),\mathbb{V}\hat{f}\left(\mathbf{x_*}\right))$, where predictive mean and variance are:
\begin{equation}
\hat{f}\left(\mathbf{x_*}\right)=\mathbf{k}_{\mathbf{x_*} \mathbf{X}}\left(\mathbf{K}_{\mathbf{X} \mathbf{X}}+\sigma_{f}^{2} \mathbf{I}\right)^{-1} \mathbf{y}, 
\label{eq:gp_mean}
\end{equation}
\begin{align}
\mathbb{V}\hat{f}\left(\mathbf{x_*}\right)=k_{\mathbf{x}_{*}\mathbf{x}_{*}}-\mathbf{k}_{\mathbf{x_*} \mathbf{X}}&\left(\mathbf{K}_{\mathbf{X} \mathbf{X}}+\sigma_{f}^{2} \mathbf{I}\right)^{-1} \mathbf{k}_{\mathbf{x_*} \mathbf{X}}^{\top}.
\label{eq:gp_var}
\end{align}
$\mathbf{k}_{\mathbf{x_*} \mathbf{X}}$ is the vector of covariances between the $J$ input points and the testing point. $k_{\mathbf{x}_{*}\mathbf{x}_{*}}$ is the covariance function value of the testing point.

Let us now briefly introduce the so-called reverting GP distance field originally presented in \cite{gentil_accurate_2023}.
Consider a surface $\mathcal{S}$ in a Euclidean space $\mathbb{R}^D$, and $\mathbf{X}=\{\abscissavec_j\}_{j=1}^{J} \in \mathbb{R}^{D}$ a set of discrete observations of $S$ as training points.
By modelling the occupancy $o(\abscissavec): \mathbb{R}^D \mapsto \mathbb{R}$ of the space with a GP as $o \sim \mathcal{GP}\left(0, k_{\abscissavec \abscissavec'}\right)$, it is possible to infer the surface occupancy $\hat{\occ}(\abscissavec_*)$ at any location in the space using Eq.~\eqref{eq:gp_mean}.
In other words, the covariance kernel $k_{\abscissavec \abscissavec'}$ is a monotonic function of the distance $\lVert \abscissavec - \abscissavec' \rVert$ between $\abscissavec$ and $\abscissavec'$.

We arbitrarily define the occupied area to be equal to 1. Therefore, $\mathbf{y}$ is equal to $\mathbf{1}$ in Eq.~\eqref{eq:gp_mean} and
\begin{align}
    \hat{\occ}(\abscissavec_*)& = \kernelvecraw{\abscissavec_*\abscissamat} \left(\kernelmatraw{\abscissamat\abscissamat} + \sigma_{\occ}^2\identitymat \right)^{-1} \mathbf{1}.
    \label{eq:occ_inference}
\end{align}
The distance field $\distfield(\abscissavec_*)$ given any location $\abscissavec_*$ is obtained by applying a \emph{reverting} function $\revert$ to the occupancy field as 
\begin{align}
    \distfield(\abscissavec_*) = \revert\left(\hat{\occ}\left(\abscissavec_*\right)\right).
    \label{eq:dist_field}
\end{align}
The reverting function depends on the chosen kernel and corresponds to the inverse operation, computing the distance as a function of the covariance between $\abscissavec$ and $\abscissavec'$:
\begin{align}
     \revert\left(\tilde{\kernelraw{}}(\lVert \abscissavec - \abscissavec' \rVert)\right) &= \lVert \abscissavec - \abscissavec' \rVert
     \\
     \text{with }\ \tilde{\kernelraw{}}(\lVert \abscissavec - \abscissavec' \rVert) &= k_{\abscissavec\abscissavec'}.
     \nonumber
\end{align}
Considering the square exponential kernel $k_{\abscissavec\abscissavec'} = \sigma^2\exp\left(-\frac{\lVert \abscissavec - \abscissavec' \rVert{}^2}{2\lengthscale^2}\right)$, substituting the reverting function of it into Eq.~\ref{eq:dist_field} gives us: 
\begin{align}
    \distfield(\abscissavec_*) = \sqrt{-2\lengthscale^2\log\left(\frac{\hat{\occ}\left(\abscissavec_*\right)}{\sigma^2}\right)}.
    \label{eq:dist_field_final}
\end{align}
This method assumes that the GP-based occupancy field in $\mathbb{R}^D$ behaves similarly to the one-noiseless-point scenario where the reverting function yields the exact distance to the data point.

\subsection{Nabla Operator for Gradient Field}
The derivative of a GP is a linear operation that produces another GP. One common solution to infer the gradient field is to have surface normals as input for GP modelling. However, this requires inverting a computational-heavy joint covariance function of $k_{\mathbf{x}\mathbf{x}'}$ with the partial derivatives of $k_{\mathbf{x}\mathbf{x}'}$ at $\mathbf{x}$ and $\mathbf{x}'$~\cite{Martens_geomprior_2017}. 
Instead, we propose to use the Nabla operator to infer the gradient along with the accurate distance field without the normal as input~\cite{le2020_GPGMaps}.
Applying this linear operator to Eq.~\ref{eq:occ_inference}, we get  
\begin{equation}
\begin{aligned}
& \nabla\hat{\occ}\left(\abscissavec_*\right)=\nabla\mathbf{k}_{\mathbf{x_*} \mathbf{X}}\left(\mathbf{K}_{\mathbf{X} \mathbf{X}}+\sigma_{\occ}^{2} \mathbf{I}\right)^{-1} \mathbf{1}, 
\end{aligned}
\label{eq:linear_operator_mean}
\end{equation}
where $\nabla\mathbf{k}_{\mathbf{x_*} \mathbf{X}}$ is the partial derivative with respect to $\mathbf{x}$~\cite{sarkka2011linear}. Taking the gradient of both sides of Eq.~\ref{eq:dist_field_final} with respect to the distance shows that the gradient of $\distfield(\abscissavec_*)$ points in the same direction as the gradient of $\hat{\occ}(\abscissavec_*)$, subject to a scaling factor, so that $\nabla \distfield(\abscissavec_*) \approx \nabla \hat{\occ}(\abscissavec_*)$. 



\begin{figure*}[ht]
    \vspace{-2ex}
    \centering
    \includegraphics[width=0.9\linewidth]{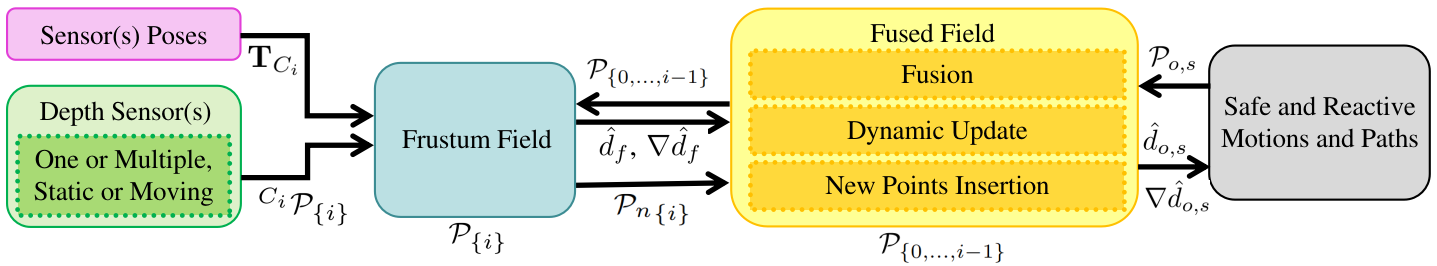}
    \caption{
    System diagram of IDMP. We first model the temporary latent Frustum Field (blue) using only $\mathcal{P}_{\{i\}}$ as training points. All prior training points $\mathcal{P}_{\{0,...,i-1\}}$ in the Fused Field (yellow) are then passed to the Frustum Field. Given the sensor pose, the Frustum Field selects from $\mathcal{P}_{\{0,...,i-1\}}$ the points that are within the frustum area $\mathcal{P}_{f\{0,...,i-1\}}$ and returns the inferred values of $\hat{d}_f$, $\nabla \hat{d}_f$ to the Fused Field. These distances and gradients are used to perform fusion and dynamic updates by updating the training points that model the Fused Field. The path planner then queries for $\hat{d}_{o,s}$, $\nabla \hat{d}_{o,s}$ in order to compute and adapt its motion plans in response to a changing map.
    }
    \label{fig:framework}
    \vspace{-2ex}
\end{figure*}

\section{Proposed Framework}\label{sec:framework}
An overview of the proposed IDMP framework is shown in Fig.~\ref{fig:framework}. Given data from one or multiple depth sensors as input, IDMP incrementally builds the \textit{Fused Field}, a continuous GP distance and gradient field for the motion planner to query and adapt its paths in response to a changing map. The measurements in the sensor frame ${C}_{{i}}$ at current time $i$ are denoted as point cloud ${ }^{{C}_{{i}}} \mathcal{P}_{\{i\}}$. This raw point cloud is projected from current sensor frame ${{C}_{{i}}}$ to world frame ${W}$ given a transformation matrix $\mathbf{T}_{{C}_{{i}}}$ yielding a point cloud $\mathcal{P}_{\{i\}}$ in the world reference frame. In the first process (blue block), we model a temporary latent GP distance and gradient field only using $\mathcal{P}_{\{i\}}$ as training points. This field only covers the region in the sensor's field of view, thus we call it \textit{Frustum Field}. This latent model allows us to query the distance and gradients with respect to the current measurements. Note that the Frustum Field is rebuilt at every frame $i$. 

The next process of IDMP (yellow block) builds and maintains our main representation, the Fused Field, generated with all prior training points $\mathcal{P}_{\{0,...,i-1\}}$. We pass $\mathcal{P}_{\{0,...,i-1\}}$ to the Frustum Field to select points only within the frustum area, denoted as $\mathcal{P}_{f\{0,...,i-1\}}$. Then we use $\mathcal{P}_{f\{0,...,i-1\}}$ to query the Frustum Field for the distance $\hat{d}_f$ and gradient $\nabla \hat{d}_f$ as per Eq.~\eqref{eq:dist_field_final}. Based on the distance and gradient inference, the fusion method adjusts the training points' positions, and the dynamic update method eliminates past training points of the objects moving in the scene. We call the intersection between $\mathcal{P}_{\{0,...,i-1\}}$ and $\mathcal{P}_{\{i\}}$ the overlapping area. Note that new measurements that are out of the range of the overlapping area, denoted as $\mathcal{P}_{n\{i\}}$, are directly merged into the Fused Field. By updating the training points, which are maintained via an Octree-based data structure, we update the Fused Field. 

For planning, let us define workspace points along the robot's links as $\mathcal{P}_{o,s}$. 
The motion planning algorithm then queries the Fused Field for $\hat{d}_{o,s}$, $\nabla \hat{d}_{o,s}$ at $\mathcal{P}_{o,s}$. These distances and gradients are used to compute a safe trajectory and can be repeatedly queried to adapt to a changing map.

\subsection{Frustum Field}\label{sec:current dis}
Every raw point transformed to the world frame $\mathcal{P}_{\{i\}}$ is used to generate the Frustum Field. Note that the Frustum Field is temporary and therefore trained and regenerated at every frame $i$ using the current measurements. As explained in Sec.~\ref{sec:background}, we use $\mathcal{P}_{\{i\}}$ as a set of discrete observations $\mathbf{X}=\{\abscissavec_j\}_{j=1}^{J}$ of surface $S$. We then model the occupancy via Eq.~\ref{eq:occ_inference} and apply the reverting function to obtain the distance field. The Frustum Field is then inferred via Eq.~\ref{eq:dist_field_final}. By using the linear operator for gradient inference, we employ Eq.~\ref{eq:linear_operator_mean} to infer the gradients along with the distance information. After the Frustum Field is modelled, we can query the distance and gradient from the points $\mathcal{P}_{f\{0,...,i-1\}}$ to pass them into the next process, the Fused Field, for the interactive updates.
\begin{figure}[ht]
    \centering
    \includegraphics[width=0.85\linewidth]{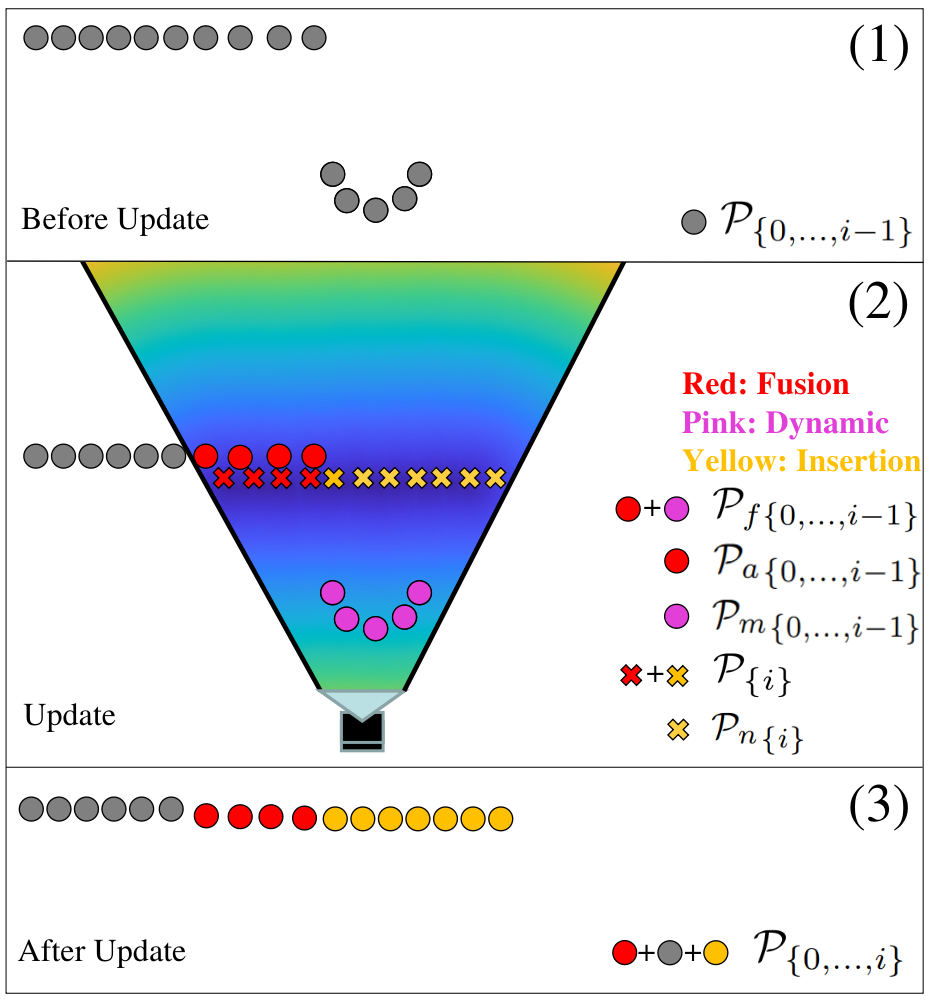}
    \caption{A simplified illustration of the proposed method. (1) Before the update, we have the prior training points $\mathcal{P}_{\{0,...,i-1\}}$ in the Fused Field. (2) The coloured area is the Frustum Field (blue is close to the surface and yellow is far from it). The current points  $\mathcal{P}_{\{i\}}$ are shown as red and yellow crosses. The prior training points within the frustum $\mathcal{P}_{f\{0,...,i-1\}}$ are marked by red and pink. Fusion, dynamic update, and insertion processes are shown in red, pink, and yellow respectively. (3) After the update, we have the updated training points $\mathcal{P}_{\{0,...,i\}}$ that model the Fused Field used by the planner.
    }


    \label{fig:fusion_dynamic}
    \vspace{-2ex}
\end{figure}
\subsection{Fused Field}
The training points are used to generate the Fused Field incrementally. Before the update at frame $i$, the Fused Field is modelled based on the prior training points $\mathcal{P}_{\{0,...,i-1\}}$ from frame $0$ to $i-1$. After fusion and dynamic update to allow merging $\mathcal{P}_{\{i\}}$ with $\mathcal{P}_{\{0,...,i-1\}}$, the Fused Field is modelled using the updated training points $\mathcal{P}_{\{0,...,i\}}$. There are three key processes performed at every frame: adjusting points for fusion, removing points for dynamic update, and inserting points for measurements in new areas. 
Fig.~\ref{fig:fusion_dynamic} illustrates the way the points are selected given the Frustum Field. The top figure shows the points $\mathcal{P}_{\{0,...,i-1\}}$ before the update given frame $i$. The middle figure shows the Frustum Field and the points involved in the three processes as described in the previous section. Red and yellow crosses are the new measurements $\mathcal{P}_{\{i\}}$ captured by the sensor. The coloured background is the Frustum Field modelled using $\mathcal{P}_{\{i\}}$. The blue area means the distance is close to the surface, and the yellow area is far away from the surface. Circles coloured in red and pink are the prior training points of the Fused Field only in the frustum, denoting as $\mathcal{P}_{f\{0,...,i-1\}}$. Note that we use the sensor pose to select $\mathcal{P}_{f\{0,...,i-1\}} \subset \mathcal{P}_{\{0,...,i-1\}}$. We then query the Frustum Field to infer $\hat{d}_f$ and gradient $\nabla \hat{d}_f$ to perform the fusion and dynamic update. An advantage of this approach is that we only query the Frustum Field once to perform the three further processes described as follows.

\subsubsection{\bfseries Fusion}
Let us denote the prior training points within the frustum as $\mathcal{P}_{a\{0,...,i-1\}} \subset \mathcal{P}_{f\{0,...,i-1\}}$ (red circles in Fig.~\ref{fig:fusion_dynamic}) and corresponding distances and gradients as $\hat{d}_{a}$, $\nabla \hat{d}_{a}$.
A threshold on the distance value $\eta$ is then used to indicate if $\mathcal{P}_{a\{0,...,i-1\}}$ are relatively close to the surface given the current Frustum Field. For each point $\mathbf{p}_{a\{0,...,i-1\}}$ in $\mathcal{P}_{a\{0,...,i-1\}}$, the proposed fusion is performed by updating its position through the distance to the surface $\hat{d}_{a}$ in the direction of the gradient $\nabla \hat{d}_{a}$ as:
\begin{equation} \label{eq:edis}
\hat{\mathbf{p}}_{a} = \mathbf{p}_{a}-\hat{d}_{a}  \nabla \hat{d}_{a}.
\end{equation}
Note that $\hat{d}_{a}$ and $\nabla \hat{d}_{a}$ are the result of simply querying the Frustum Field. No data association nor iteration is required as GP distance inference produces directly the distance to the surface and it is accurate through our reverting GP model. After the fusion, $\mathcal{P}_{a\{0,...,i-1\}}$ is replaced by $\hat{\mathcal{P}}_{a\{0,...,i-1\}}$ as the training point of the Fused Field. Note that since we have $\hat{\mathcal{P}}_{a\{0,...,i-1\}}$, we do not need to include new points in the overlapping area (red crosses) into the Fused Field. 

\subsubsection{\bfseries Dynamic Update}
We show the dynamic update in Fig.~\ref{fig:fusion_dynamic}. Dynamic objects are handled based on the information provided by the Frustum Field as well, as we only update the moving objects in the current field of view. 
Let us denote the prior training points to-be-removed as $\mathcal{P}_{m\{0,...,i-1\}}\subset \mathcal{P}_{f\{0,...,i-1\}}$ (pink circles). Note that in this case and without loss of generality, we can capture measurements behind $\mathcal{P}_{f\{0,...,i-1\}}$ represents that the points $\mathcal{P}_{f\{0,...,i-1\}}$ have been removed and are not in the scene anymore. The Frustum Field distance values $\hat{d}_{m}$ for these points are bigger than $\eta$ to indicate they are relatively far away from the current captured measurements, which means the object has moved. We then eliminate from the training set the removed points $\mathcal{P}_{m\{0,...,i-1\}}$ of the Fused Field. Note that in some cases, the points to adjust and remove are very close to each other and may not be separated explicitly. 

\subsubsection{\bfseries New Points Insertion}Let us denote the points in $\mathcal{P}_{\{i\}}$ that are outside of the overlapping area as $\mathcal{P}_{n\{i\}}$ (yellow crosses). These points are directly added to the Fused Field training set. Points outside the overlapping area in the Fused Field remain the same (grey circles). Following our illustration, at the bottom figure in Fig.~\ref{fig:fusion_dynamic} after the update, we use the updated training points $\mathcal{P}_{\{0,...,i\}}$ to model the Fused Field and have it ready for the motion planner query. 

We only use one Frustum Field for all interactive updates. The Frustum Field allows to perform the fusion in just one-step and to clear the removed object effectively.

\subsection{Reactive Planning}\label{reactive_planning_method}
Due to the differentiability \edit{and efficiency} of the Fused Field, our framework naturally accommodates \edit{reactive planning. For example, a gradient-based motion planner} can query $\hat{d}_{o}$ and $\nabla \hat{d}_{o}$ and utilise these to adapt its motion plans in response to a changing map. A key advantage of our framework is that we can use the gradient inference in Eq.~\ref{eq:linear_operator_mean} to compute the gradients analytically, whereas conventional approaches rely on discrete grid-based approximation~\cite{zucker2013chomp}. This results in faster and more accurate computation of gradients which is crucial for close-proximity human-robot collaborative tasks where fast replanning is required. 

\edit{In addition, 
our framework naturally accommodates fast local reactive methods. Consider the current point $\mathbf{x}_s$ and a goal point $\mathbf{x}_g$. Given the queried distance and gradient of the current point as $\hat{d}_s$ and $\nabla\hat{d}_s$ from our IDMP Fused Field, we can formulate a repulsive vector $\mathbf{v}_{rep} = \nabla \hat{d}_s$ and an attractive vector $\mathbf{v}_{att} = \mathbf{x}_g-\mathbf{x}_s$. Note that both vectors need to be normalised. The repulsive behaviour is defined by a function $w(\hat{d}_s)$ dependent on the distance. We then can formulate the resulting vector by combining the repulsive vector and attractive vector using $w(\hat{d}_s)$ as in~\cite{khatib_real-time_1986},
\begin{equation}
    \mathbf{v}_{res} = w(\hat{d}_s)\mathbf{v}_{rep} + (1-w(\hat{d}_s))\mathbf{v}_{att}.
    \label{reactive_formulation}
\end{equation}
By efficiently computing Eq.~\ref{reactive_formulation} at each timestep, a robot has the ability to follow the resulting vector to reactively avoid moving objects present in the scene in an online manner. 
}


\section{Evaluation}\label{sec:evaluation}

To evaluate the proposed IDMP framework, we a) quantitatively compare the mapping performance for both static and dynamic scenes with other frameworks, and b) qualitatively demonstrate online motion planning for a dynamic scene.

\subsection{Distance Field Mapping}

We compare IDMP to the two state-of-the-art algorithms: FIESTA \cite{han_fiesta_2019} and Voxblox \cite{oleynikova_voxblox_2017}. Both are designed to compute the distance field observed by a 3D sensor online. One key difference from both approaches to IDMP is the need to specify a fixed voxel size upfront which directly influences performance. In contrast, IDMP uses a continuous function to represent the field that can be queried at any point in space. We also demonstrate IDMP's ability to handle dynamic scenes with unknown moving objects and compare the performance to FIESTA/Voxblox.

The evaluation is conducted on two distinct datasets: a) the widely used Cow \& Lady dataset~\cite{oleynikova_voxblox_2017}, and b) a custom synthetic dataset featuring a dynamically moving ball on a table. These datasets are chosen to assess IDMP's accuracy in static and dynamic scenarios relative to current state-of-the-art algorithms. 
%
The Cow \& Lady dataset features complex surface geometries of static objects in a room recorded by a moving Kinect-1 RGB-D camera whose pose is tracked by a vicon system. The ground truth surface reconstruction is obtained by a Leica TPS-M50 laser scanner. 
%
The second dataset is used for evaluating a dynamic scene. It is generated using Gazebo and features a ball rolling on a table observed by a simulated RGB-D camera. The ground truth for distances and gradients are obtained from Gazebo.

The experiments are conducted on a laptop equipped with an AMD Ryzen 7 4800u CPU, 16GB RAM and no GPU. All three algorithms are allowed to multithread on up to 16 threads. Each algorithm is allowed to run as a ROS node with only the playback of the dataset (rosbag) running in the background. Original code implementations for Fiesta\footnote{\url{https://github.com/HKUST-Aerial-Robotics/FIESTA}} and Voxblox\footnote{\url{https://github.com/ethz-asl/voxblox}} were used for the experiments with default parameters.
Root Mean Squared Error (RMSE) is used to evaluate the distance field performance. For gradients, we use
\begin{equation} \label{eq:cos_sim}
    \text{cosine\_similarity}(\mathbf{a}, \mathbf{b}) = \frac{\mathbf{a} \cdot \mathbf{b}}{\|\mathbf{a}\| \cdot \|\mathbf{b}\|}
\end{equation}
to measure the difference between two gradient vectors $\mathbf{a}$ and $\mathbf{b}$ in terms of their direction.



\subsection{Mapping Results}

\subsubsection{Static Scene}

\begin{figure}[hb]
    \centering
    \includegraphics[width=0.95\linewidth]{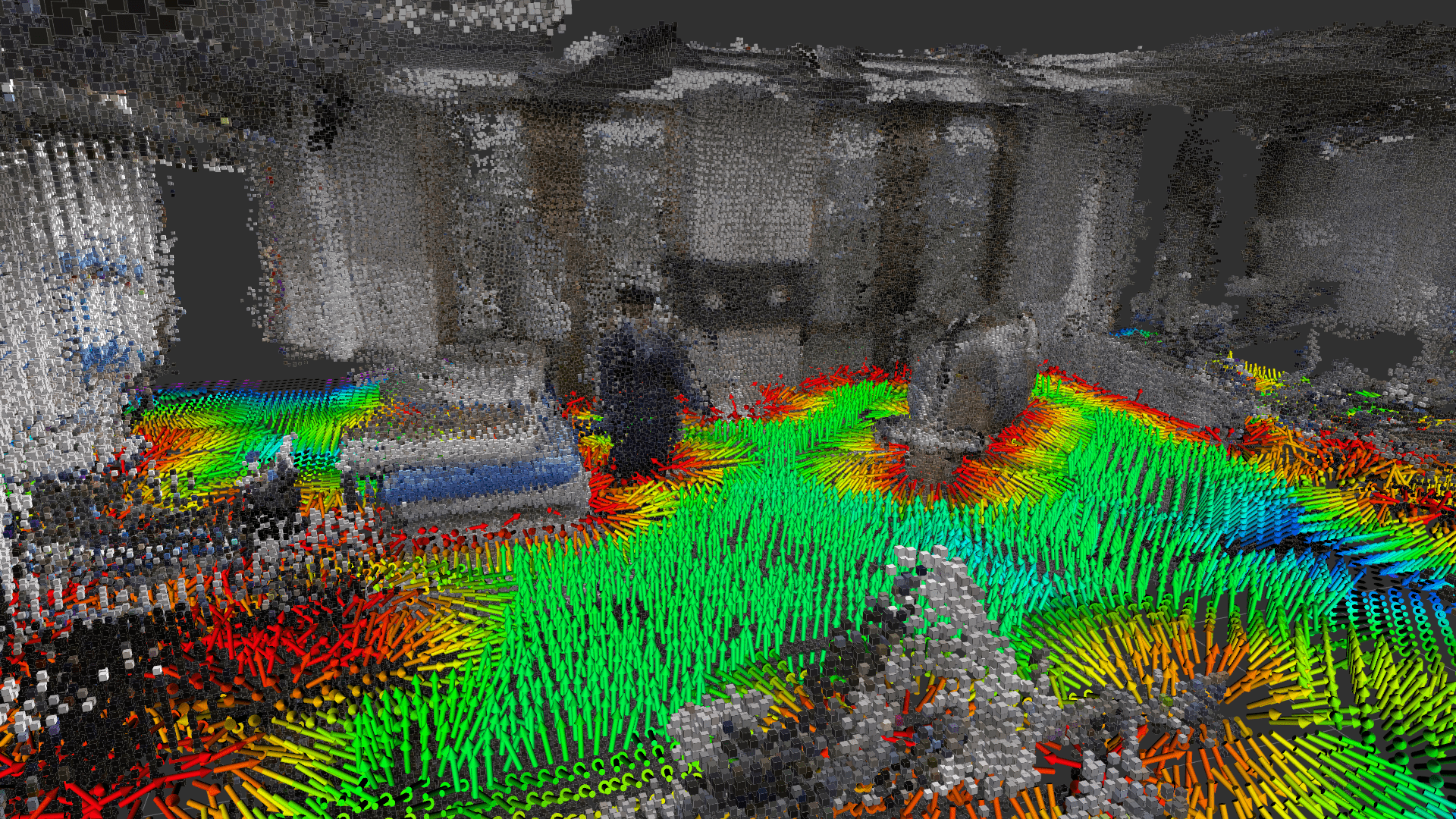}
    \caption{Example of an IDMP query on a horizontal plane for the Cow \& Lady dataset. The colour of the arrows represents the distance to the nearest object. The arrows are normalised gradients that point in the direction away from the nearest object. The updated training points are coloured using RGB data from the camera.}
    \label{fig:dist_field_example_cow}
\end{figure}

Fig.~\ref{fig:dist_field_example_HRC} and Fig.~\ref{fig:dist_field_example_cow} present qualitative results where IDMP queries are visualised. For every point in space, a distance (colour) and direction (arrow) away from the closest obstacle can be computed online and on-demand. For both examples, the queries are taken using a regular grid on a plane parallel to the table and floor, respectively. Note that because of the continuous nature of IDMP, the query points can be located anywhere in space and are neither constrained to be on a plane or a regular grid nor bound to a specific spatial resolution.

\begin{figure}[ht]
    \vspace{-2ex}
	\centering
    \includegraphics[width=0.95\linewidth]{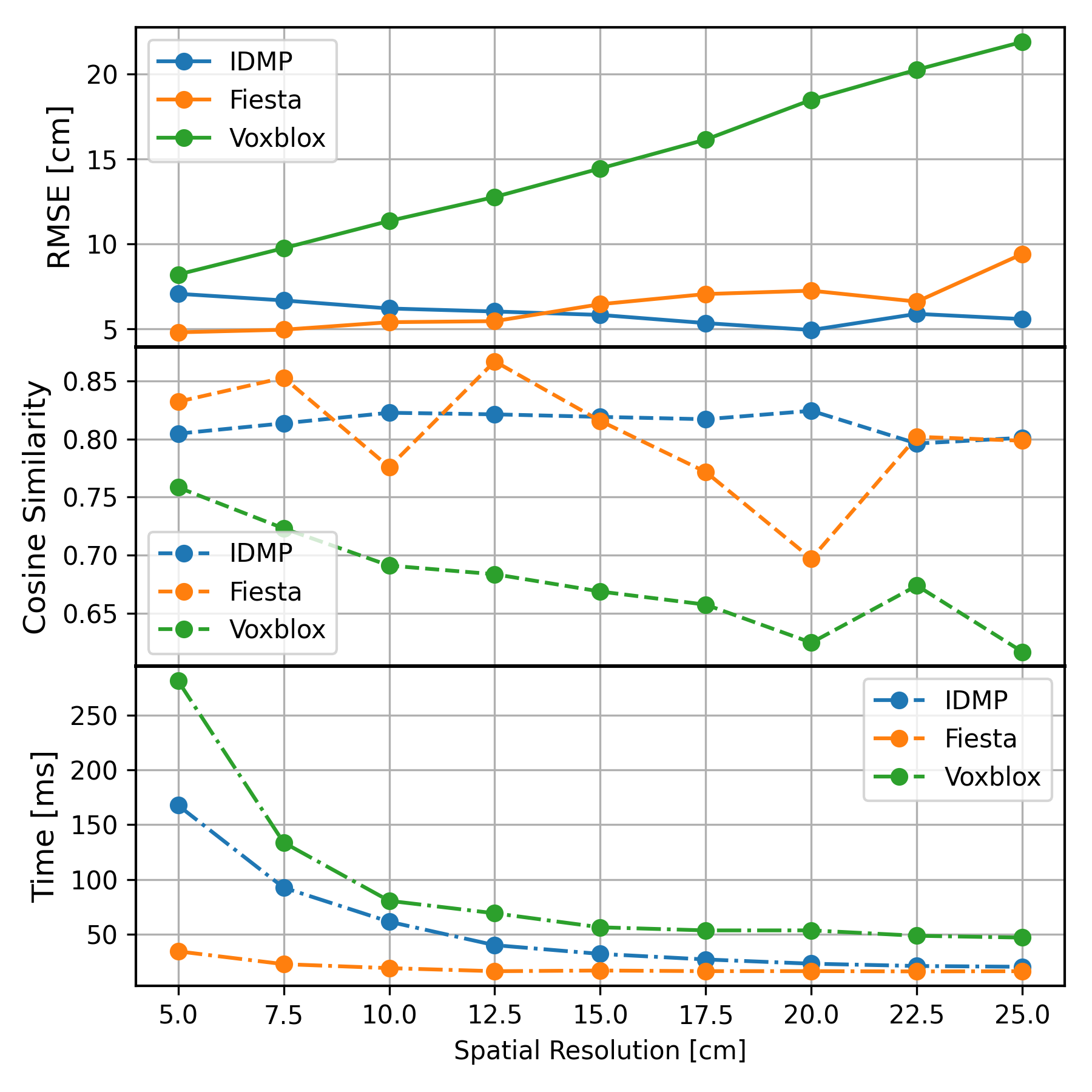}
	\caption{\edit{Quantitative evaluation of distance RMSE (lower is better), gradients cosine similarity (higher is better) and computation time (lower is better) using the Cow \& Lady dataset for IDMP, FIESTA and Voxblox.} 
    }
	\label{fig:cow_dist_grad}
	\vspace{-2ex}
\end{figure}



To compare IDMP to Voxblox and FIESTA, we compute the RMSE for each framework while varying the \textit{spatial resolution}. For Voxblox/FIESTA, this corresponds to the voxel size. For IDMP we \edit{vary the resolution of the training points used to update the Fused Field}. Figure \ref{fig:cow_dist_grad} shows our experimental results. \edit{It shows the distance and gradient accuracy, and the overall time needed to process a single frame.} 

The accuracy for Voxblox and FIESTA decreases with increasing spatial resolution. In contrast, IDMP's accuracy remains constant, \edit{outperforming FIESTA with resolutions higher than 15cm. Note that even though we increase the resolution of our training points, IDMP generates a continuous distance field, that can be queried at any point in space independent of the spatial resolution. Voxblox and FIESTA only generate a discrete distance field in the predefined spatial resolution and where data was observed.} 

The evaluation of gradients shows the mean of the cosine similarities between the computed gradients and the ground truth. Both IDMP and FIESTA perform better than Voxblox for all spatial resolutions. FIESTA shows slightly better performance than IDMP in some cases; however, at the cost of higher variability. 

\edit{The last part shows the average computation time for each algorithm to process a frame. For IDMP this includes generating the Frustum Field, performing the fusion, updating dynamic objects and generating the Fused Field. For Fiesta and Voxblox this includes the ESDF/TSDF generation, integration and raycasting. IDMP outperforms Voxblox for every spatial resolution while reaching similar computation times as FIESTA for resolutions larger than $15$ cm.

Regarding query times, it should be noted that while Voxblox and Fiesta have constant distance and gradient query times, IDMP has a small overhead time ($2\mu$s per point). However, the benefit of our method is that queries can be made at any arbitrary resolution post-mapping due to our continuous representation.
}

\subsubsection{Dynamic Scene}
\begin{figure*}[ht]
    \centering
    \begin{subfigure}{0.23\linewidth}
		\includegraphics[width=\linewidth]{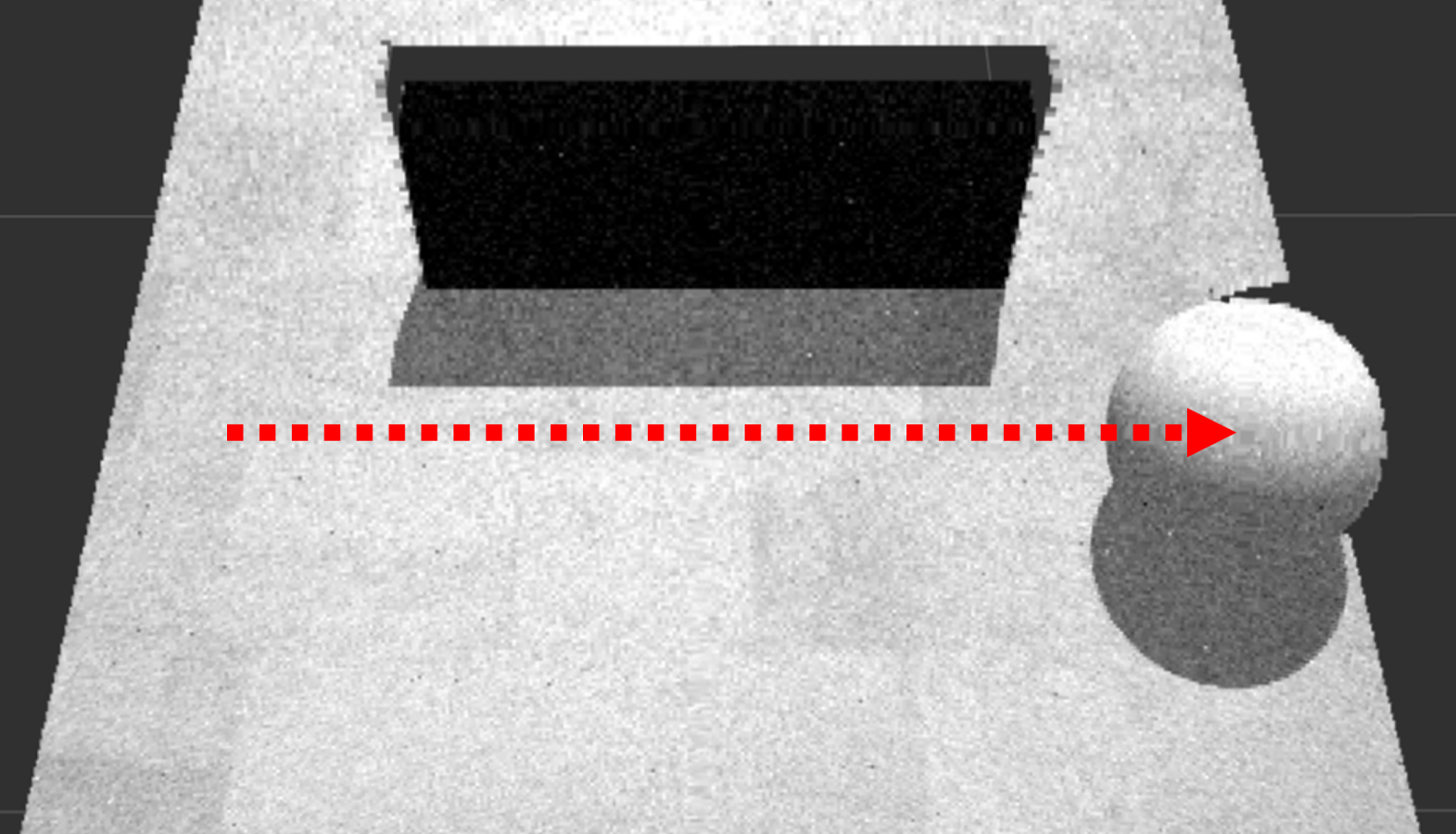} 
		\caption{Simulated scene}
		\label{fig:ball_scene}
    \end{subfigure}
    \hfill
    \begin{subfigure}{0.23\linewidth}
		\includegraphics[width=\linewidth]{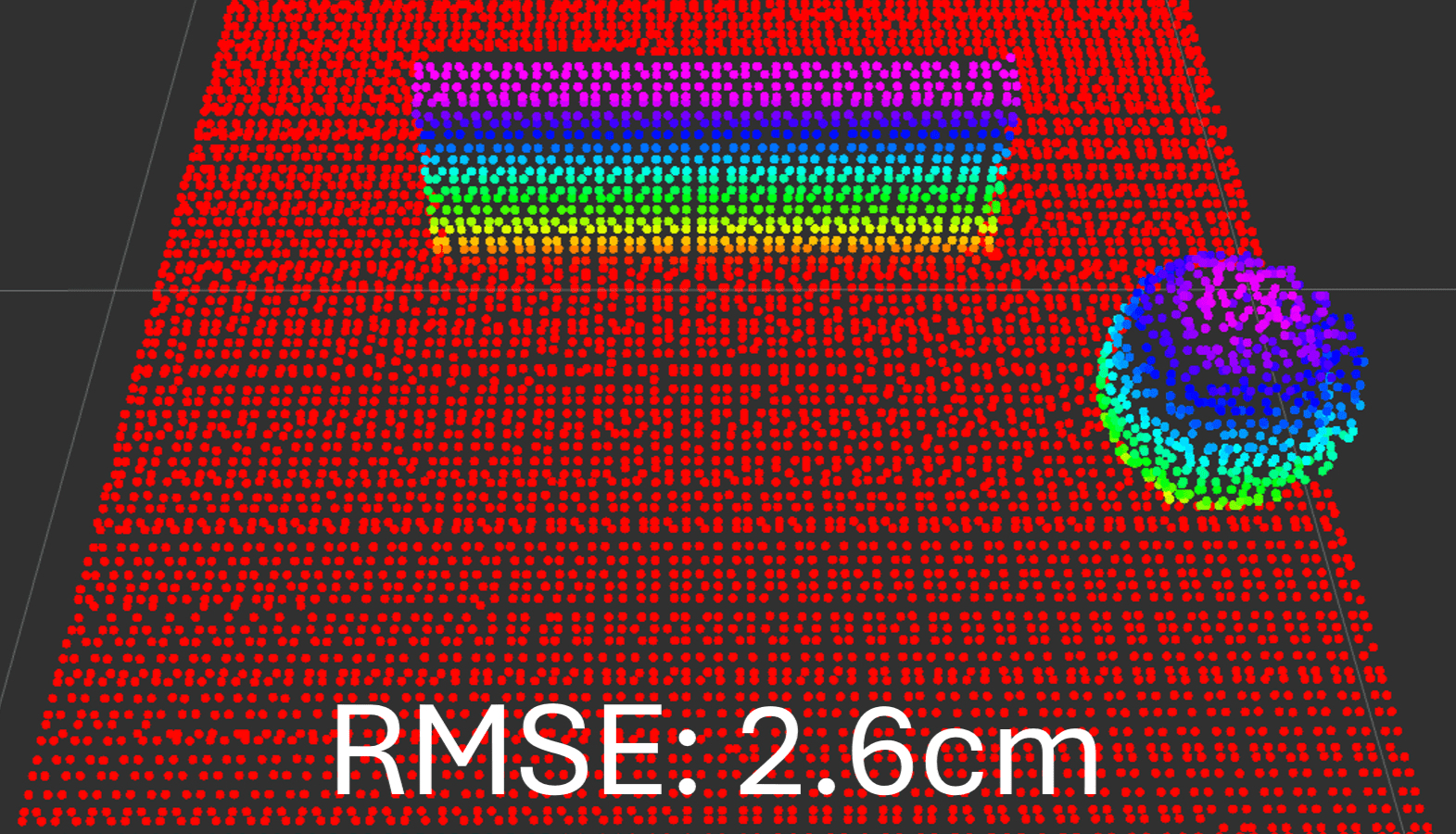} 
		\caption{IDMP}
		\label{fig:ball_IDMP}
    \end{subfigure}
    \hfill
    \begin{subfigure}{0.23\linewidth}
		\includegraphics[width=\linewidth]{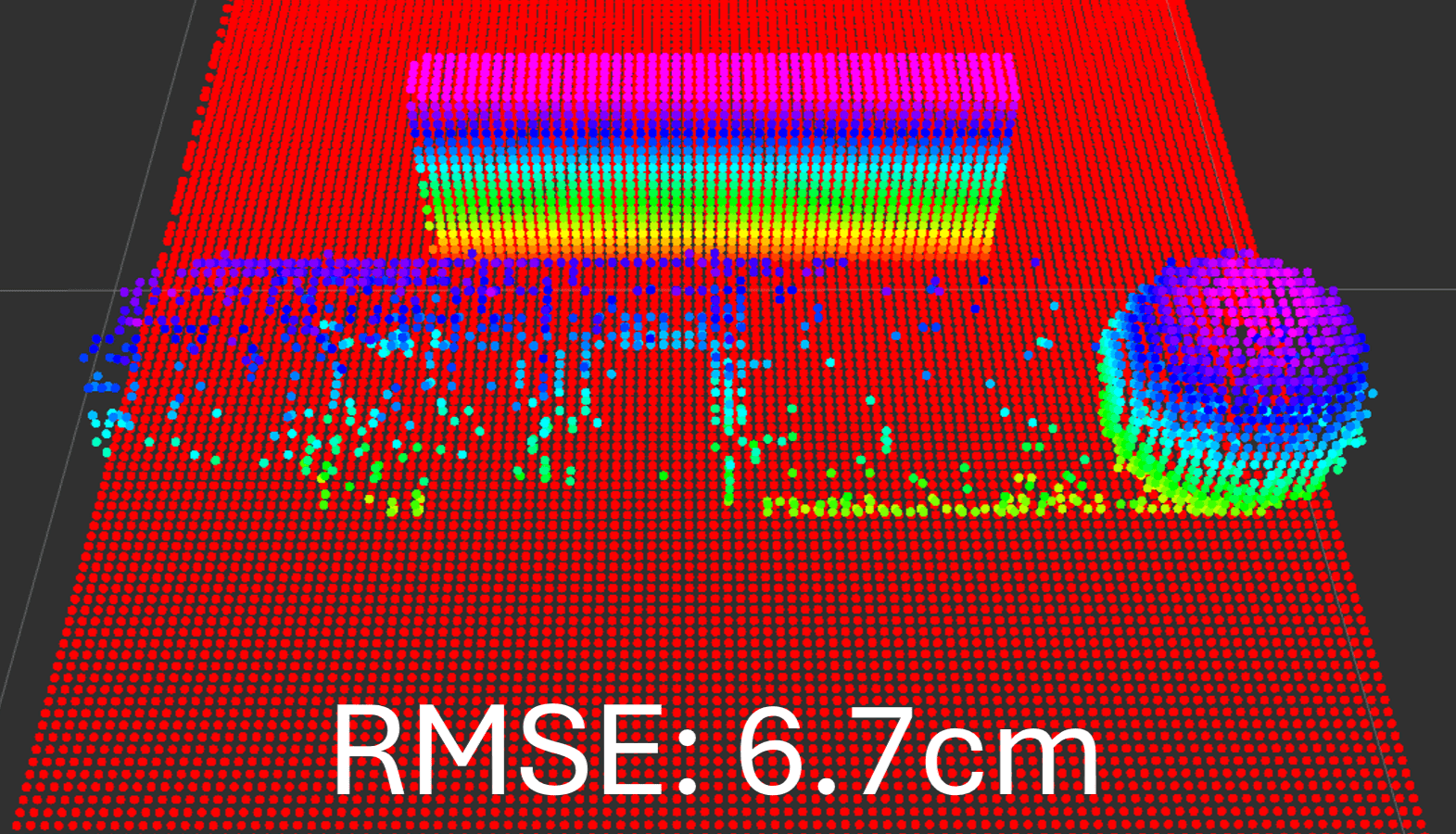} 
		\caption{FIESTA}
		\label{fig:ball_fiesta}
    \end{subfigure}
    \hfill
    \begin{subfigure}{0.23\linewidth}
		\includegraphics[width=\linewidth]{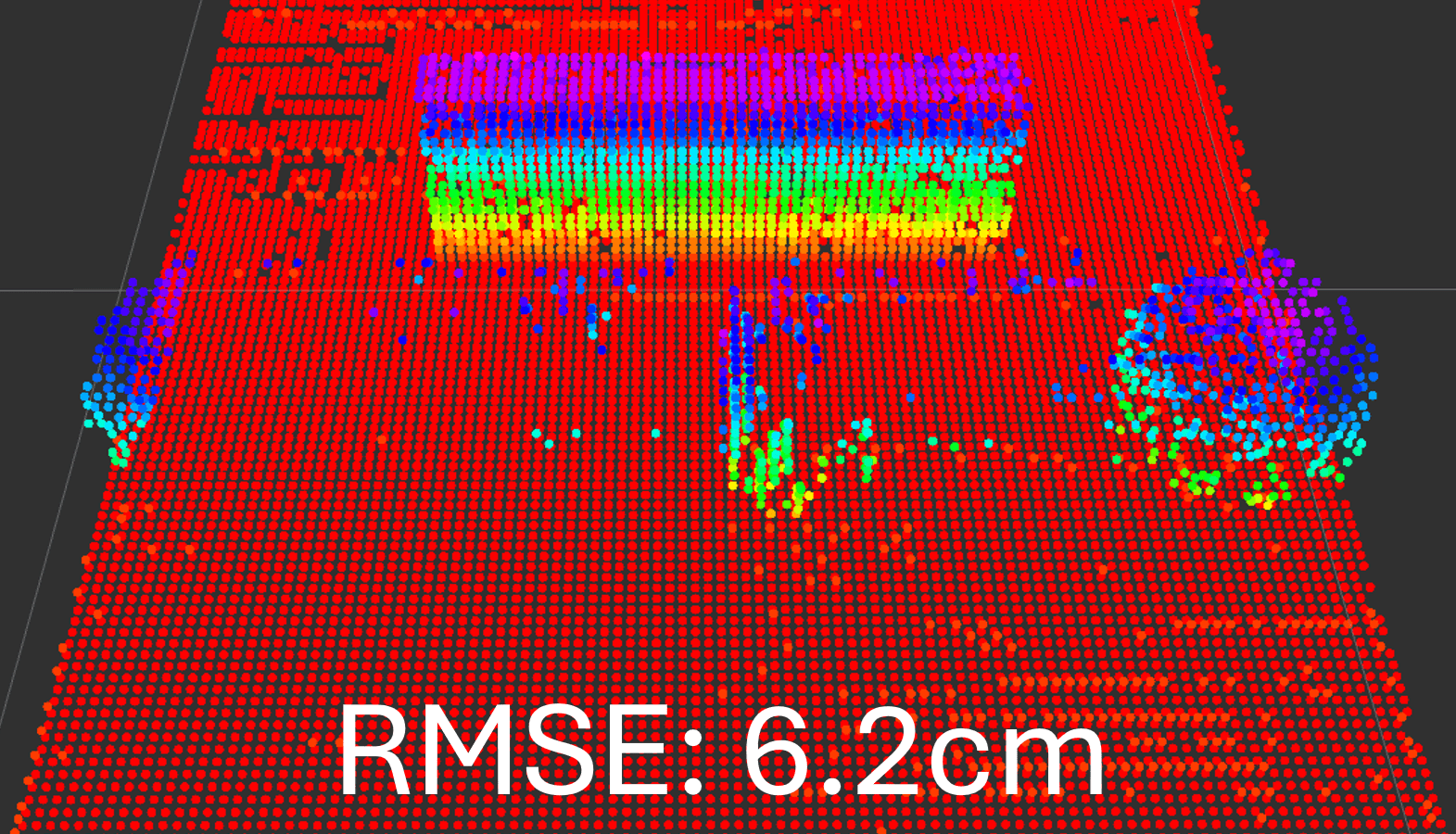} 
		\caption{Voxblox}
		\label{fig:ball_vb}
    \end{subfigure}
    \caption{Comparison of IDMP, FIESTA and Voxblox for a dynamic scene. (a) Shows the stimulation setup, and (b) to (d) show the fused surface points at the end of the sequence. \edit{Colours indicate the point's height with a spatial resolution of 1 cm. IDMP handles appropriately moving objects while FIESTA and Voxblox keep artefacts from older frames.}}
    \label{fig:ball_scenes}
    \vspace{-2ex}
\end{figure*}

In this experiment, we evaluate how each framework handles dynamic scenes with moving objects. Fig.~\ref{fig:ball_scene} shows the Gazebo scene with a ball rolling on the table from left to right. The last three subfigures show the surface points at the end of the run for the three frameworks. Both FIESTA and Voxblox show artefacts due to the progressive weighting and integration, while IDMP does not. IDMP results in the best RMSE of $2.6$ cm which is more than a factor of $2$ better than FIESTA/Voxblox. Updating the GP for a scene of this size takes on average $50$ ms per frame while the query step takes $2~ \mu$s per point. Only the time for processing the frame is dependent on the size of the scene, whereas the time needed for the query remains constant.

\subsection{Reactive Planning in Dynamic Scenes}\label{reactive_planning}
\begin{figure}[ht]
    \centering
    \begin{subfigure}{0.49\linewidth}
		\includegraphics[width=\linewidth]{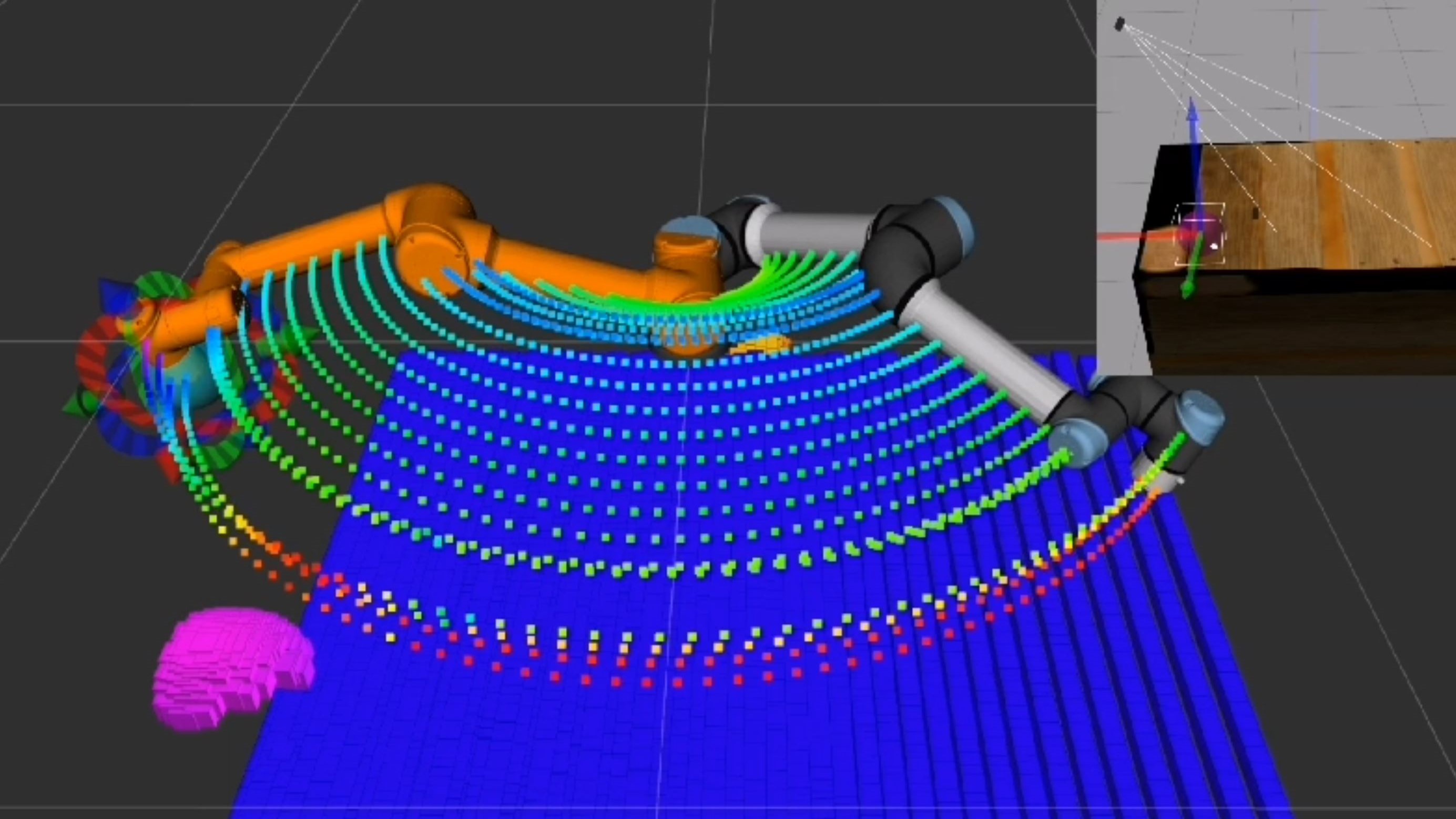} 
		\caption{Original planned trajectory (no obstacle encountered).}
		\label{fig:chomp_example_a}
    \end{subfigure}
    \hfill
    \begin{subfigure}{0.49\linewidth}
		\includegraphics[width=\linewidth]{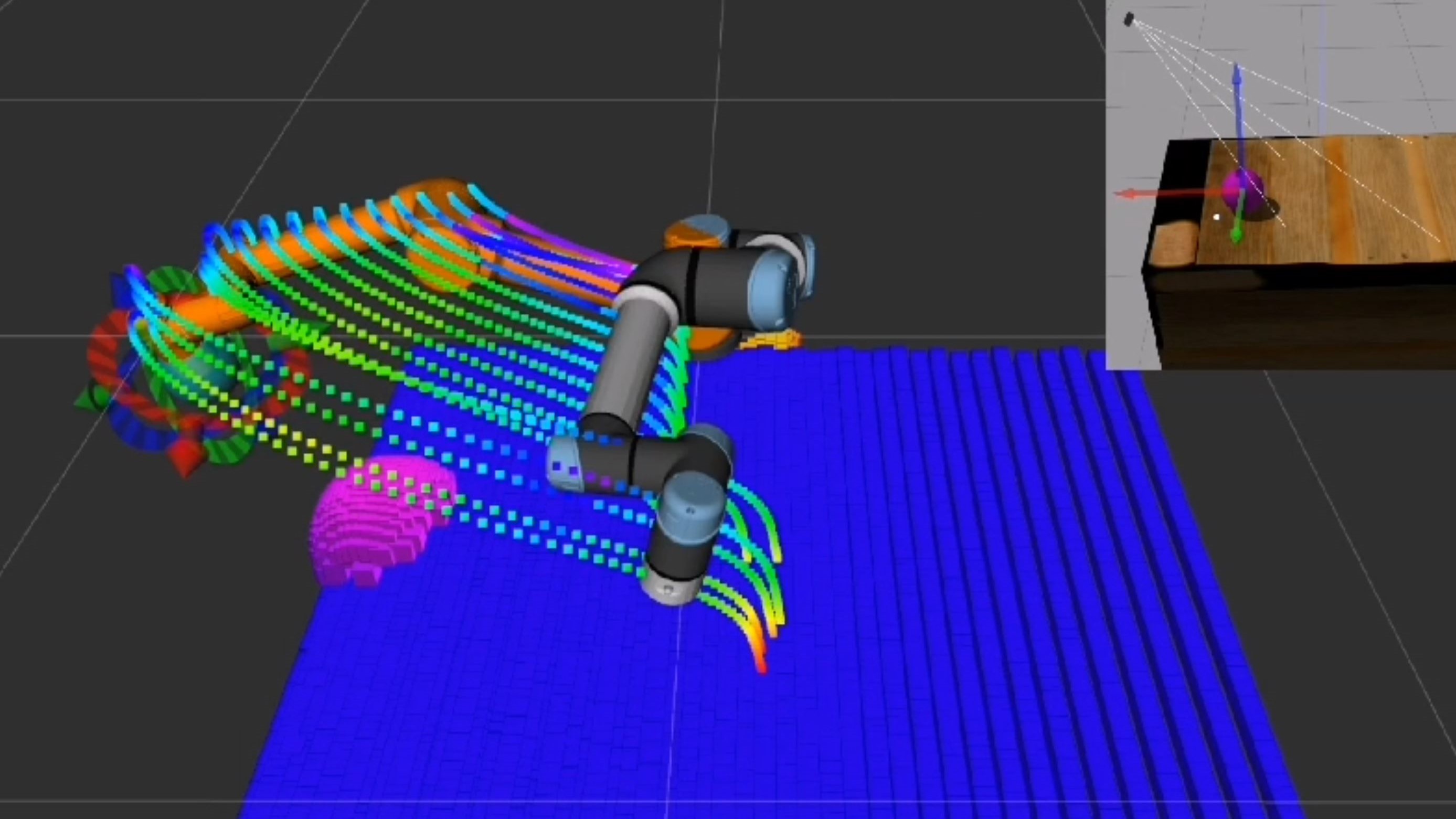} 
		\caption{Modified trajectory due to dynamic obstacle.}
		\label{fig:chomp_example_b}
    \end{subfigure}
    \caption{Re-planning in the presence of a dynamic obstacle. CHOMP approximates the manipulator via $29$ spheres. For the trajectory shown, $110$ points per sphere are used, resulting in $29\times110=3190$ query points.}
    \label{fig:chomp_example}
    \vspace{-2ex}
\end{figure}

In this experiment, we demonstrate that our method is able to facilitate safe \edit{reactive motion and path} planning of a robot arm in the presence of dynamic obstacles, which is useful for close-proximity human-robot collaboration tasks. To simulate such a scenario, we task the robot with moving to a goal pose while avoiding a moving ball object, see Fig.~\ref{fig:chomp_example}.

Trajectory optimisers are well suited to such scenarios due to their fast computation and smooth trajectory properties. We utilise CHOMP~\cite{zucker2013chomp}, a gradient-based trajectory optimisation method that iteratively perturbs a given initial trajectory away from obstacles. To do so, it utilises the gradient field to minimise its optimisation objective, which consists of trajectory smoothness and obstacle avoidance terms. Due to a) the ability of our framework to compute updated gradients analytically and b) the need to only query the points $\mathcal{P}_{o}$ the motion planner requires, we can continuously replan online. The computation time for the query shown in Fig.~\ref{fig:chomp_example} is approximately $6.4$ ms ($3190$ points at $2~\mu $s point).

\begin{figure}[hb]
    \vspace{-2ex}
    \centering
    \includegraphics[width=0.8\linewidth]{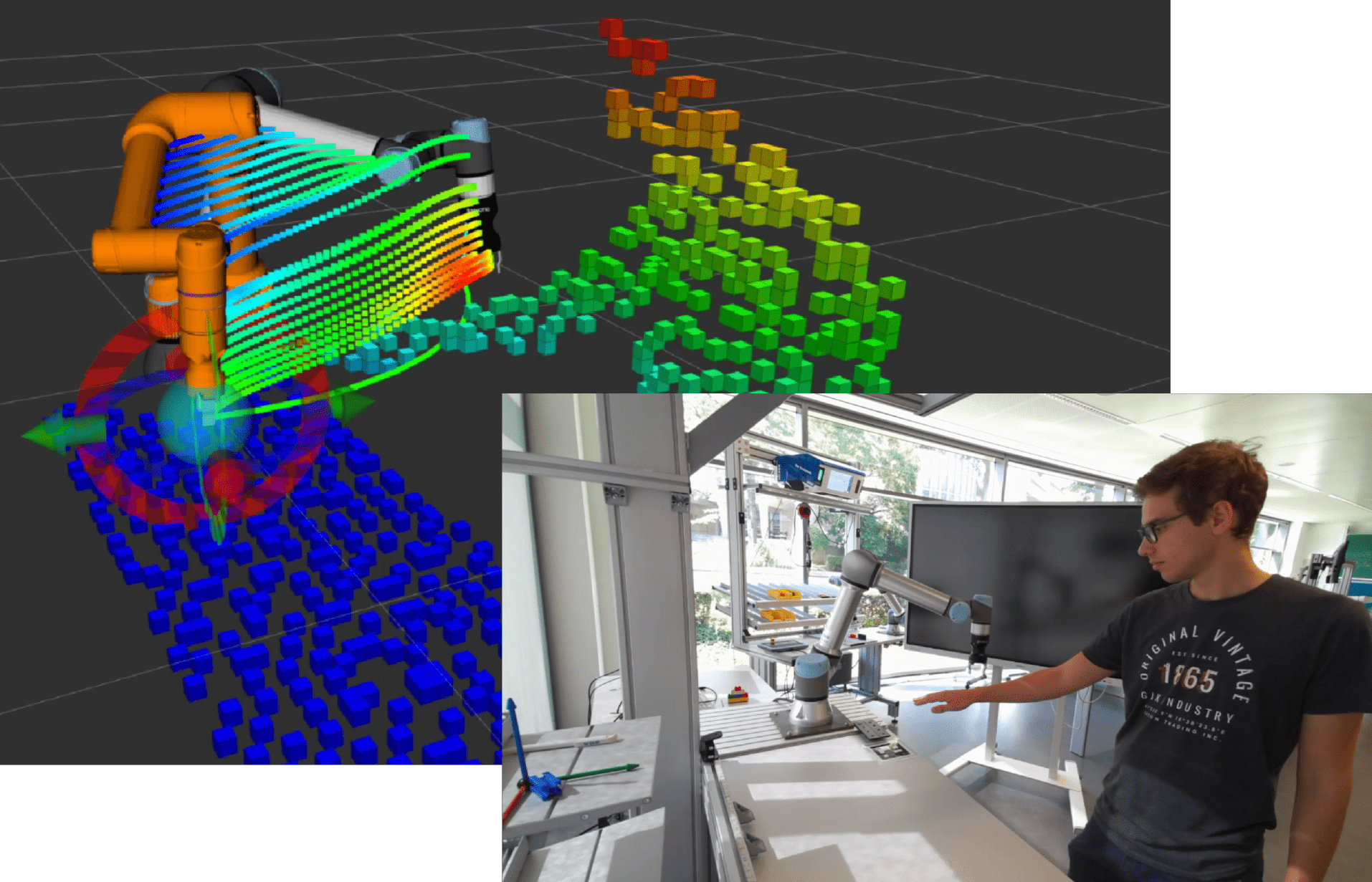}
    \caption{\edit{Real-world experiment of IDMP with CHOMP. Re-planning in the presence of a human arm.}}
    \label{fig:chomp_real}
    \vspace{-2ex}
\end{figure}

\begin{figure}[htb]
    \vspace{-2ex}
    \centering
    \includegraphics[width=0.8\linewidth]{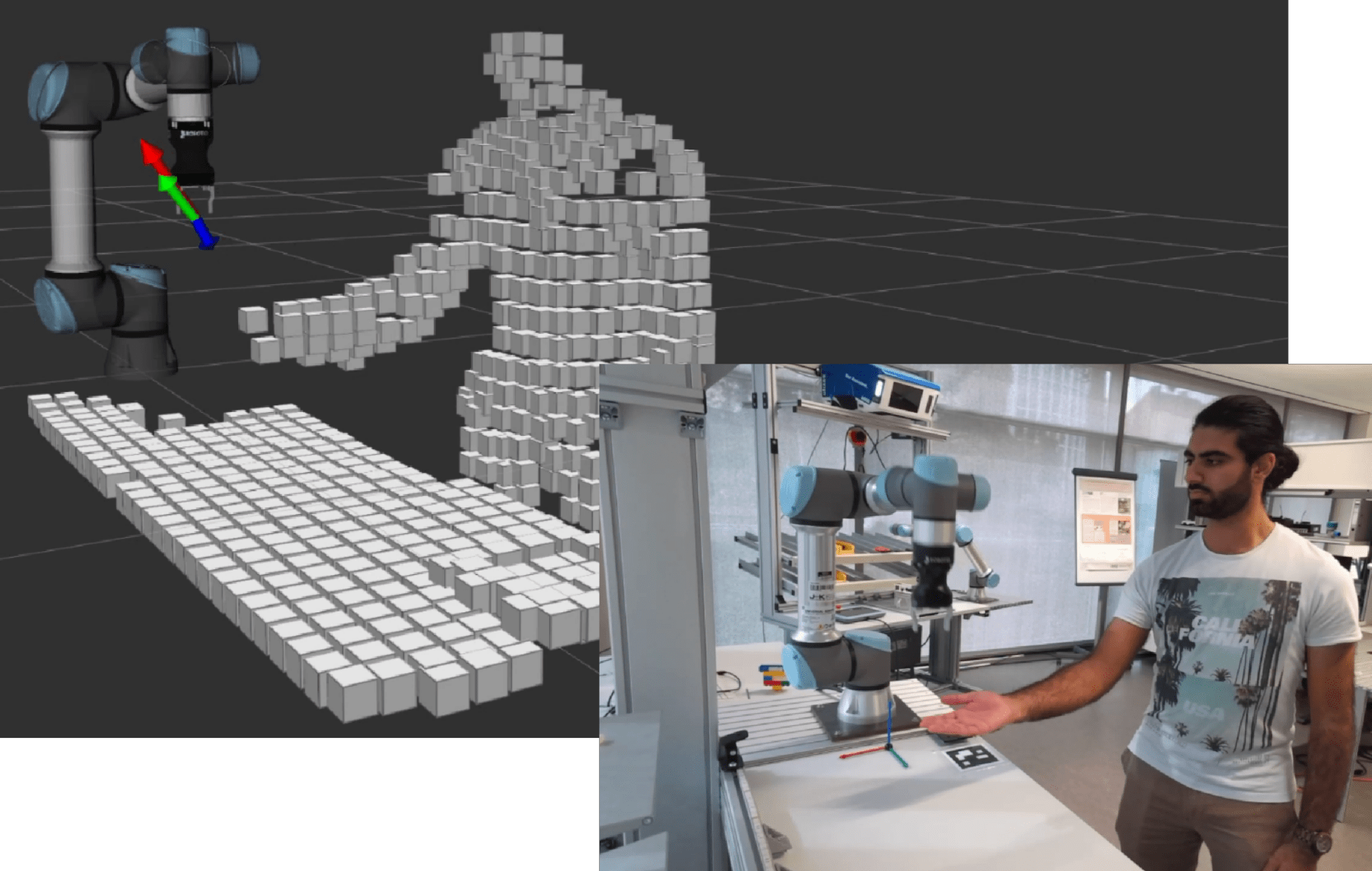}
    \caption{\edit{IDMP and local reactive obstacle avoidance.}}
    \label{fig:rep_vector}
    \vspace{-2ex}
\end{figure}

\edit{We additionally validate IDMP in combination with CHOMP in a real-world scenario where a human moves into the robot's workspace mid-way during its trajectory (see Fig.~\ref{fig:chomp_real}). Furthermore, we demonstrate the proposed local reactive method in a similar scenario. Fig.~\ref{fig:rep_vector} shows the repulsive vector in red, the attractive vector in blue and the resulting vector in green. The robot arm follows the resulting vector to quickly and safely avoid the human's moving arm. Because the repulsive vector only uses the information at the robots current position, querying the Fused Field takes $2 ~\mu $s, enabling real-time obstacle avoidance.
}


\section{Conclusion}\label{sec:conclusion}
In this letter, we presented an efficient interactive mapping and planning framework named IDMP. The simple and elegant dynamic updates and fusion are performed by querying distances and gradients from a temporary latent GP distance field generated with only the points of the current frame. Our framework can run online on a modest PC without a GPU and handles dynamic objects effectively. The fused representation is continuous enabling queries of the Euclidean distance with gradient at an arbitrary spatial resolution. This property enables seamless integration with a gradient-based motion planning framework for HRC applications. Going forward, we aim to further explore IDMP's capabilities, such as multi-resolution sampling and mapping uncertainty, for enhancing motion planning, particularly in HRC applications.


\section{Acknowledgement}\label{sec:acknowledgement}
The authors would like to thank Cedric Le Gentil for useful discussions and code collaboration.

%
%

\ifCLASSOPTIONcaptionsoff
  \newpage
\fi

\begin{acronym}[AAAAAA]
    \acro{GP}{Gaussian Process }
    \acro{GPIS}{Gaussian Process Implicit Surfaces }
    \acro{CHOMP}{Covariant Hamiltonian Optimization for Motion Planning}
    \acro{EDF}{Euclidean Distance Field }
    \acro{EDFs}{Euclidean Distance Fields }
    \acro{TSDF}{Truncated Signed Distance Field }
    \acro{ESDF}{Euclidean Signed Distance Field }

\end{acronym}

\bibliographystyle{IEEEtran}
\bibliography{reference}
\end{document}